\documentclass[11pt,letterpaper]{article}
\usepackage{naaclhlt2016}
\usepackage{times}
\usepackage{latexsym}

\usepackage{helvet}
\usepackage{courier}
\usepackage{graphicx}
\usepackage{amsmath,amsfonts,amsthm}
\usepackage{epstopdf}
\usepackage{url}
\naaclfinalcopy 
\def\naaclpaperid{***} 

\title{A C-LSTM Neural Network for Text Classification}

\date{}
\author{{Chunting Zhou}$^{1}$, Chonglin Sun$^{2}$, Zhiyuan Liu$^{3}$, Francis C.M. Lau$^{1}$\\
Department of Computer Science, The University of Hong Kong$^{1}$\\
School of Innovation Experiment, Dalian University of Technology$^{2}$\\
Department of Computer Science and Technology, Tsinghua University, Beijing$^{3}$\\
}

\begin{document}

\maketitle

\begin{abstract}
Neural network models have been demonstrated to be capable of achieving
remarkable performance in sentence and document modeling.
Convolutional neural network (CNN) and recurrent neural network
(RNN) are two mainstream architectures for such modeling tasks, which
adopt totally different ways of understanding natural languages.
In this work, we combine the strengths of both architectures
and propose a novel and unified model called C-LSTM for sentence
representation and text classification.
C-LSTM utilizes CNN to extract a sequence of higher-level phrase
representations, and are fed into a long short-term
memory recurrent neural network (LSTM) to obtain the sentence
representation.
C-LSTM is able to capture both local features of phrases as
well as global and temporal sentence semantics.
We evaluate the proposed architecture on sentiment classification and
question classification tasks.
The experimental results show that the C-LSTM outperforms both CNN and LSTM and can achieve excellent performance on these tasks.

\end{abstract}

\section{Introduction}
As one of the core steps in NLP, sentence modeling aims at representing
sentences as meaningful features for 
tasks such as sentiment classification.
Traditional sentence modeling uses the bag-of-words model
which often suffers from the curse of dimensionality; others use composition
based methods instead, e.g., an algebraic operation over semantic word vectors
to produce the semantic sentence vector. However, such methods may not
perform well due to the loss of word order information.
More recent models for distributed sentence representation fall into two
categories according to the form of input sentence: sequence-based
models and tree-structured models.
Sequence-based models construct sentence representations from word
sequences by taking in account the relationship between successive
words~\cite{effective}.
Tree-structured models 
treat each word token as a node in a syntactic parse tree and learn sentence
representations from leaves to the root in a recursive
manner~\cite{socher2013}.\\
\indent Convolutional neural networks (CNNs) and recurrent neural
networks (RNNs) have emerged as two widely used architectures and are
often combined with sequence-based or tree-structured
models~\cite{tai2015,tao,tang,kim,dcnn,mou}.\\
\indent Owing to the capability of capturing local correlations of
spatial or temporal structures, CNNs have achieved top performance in
computer vision, speech recognition and NLP.
For sentence modeling, CNNs perform excellently in extracting n-gram
features at different positions of a sentence through convolutional
filters, and can learn short and long-range relations through pooling
operations.
CNNs have been successfully combined with both sequence-based
model~\cite{modelling,dcnn} and tree-structured model~\cite{mou} in
sentence modeling.\\
\indent The other popular neural network architecture -- RNN -- is able
to handle sequences of any length and capture long-term dependencies.
To avoid the problem of gradient exploding or vanishing in the standard
RNN, Long Short-term Memory RNN (LSTM)~\cite{lstm} and other
variants~\cite{gru} were designed for better remembering and
memory accesses.
Along with the sequence-based~\cite{tang} or the
tree-structured~\cite{tai2015} models, RNNs have achieved remarkable
results in sentence or document modeling.\\
\indent To conclude, CNN is able to learn local response from temporal or spatial data but lacks the ability of learning sequential correlations; on the other hand, RNN is specilized for sequential modelling but unable to extract features in a parallel way.
It has been shown that higher-level modeling of $x_t$ can help to disentangle underlying factors of variation within the input, which should then make it easier to learn temporal structure between successive time steps~\cite{evidence}. For example, Sainath et al.~\cite{saina} have obtained respectable improvements in WER by learning a deep LSTM from multi-scale  inputs.
We explore training the LSTM model directly from sequences of higher-level representaions while preserving the sequence order of these representaions.
In this paper, we introduce a new architecture short for C-LSTM by combining CNN and LSTM to model sentences.
To benefit from the advantages of both CNN and RNN, we design a simple end-to-end, unified architecture by feeding the output of a one-layer CNN into LSTM.
The CNN is constructed on top of the pre-trained word vectors from massive unlabeled text data to learn higher-level representions of n-grams.
Then to learn sequential correlations from higher-level suqence representations, the feature maps of CNN are organized as sequential window features to serve as the input of LSTM.
In this way, instead of constructing LSTM directly from the input sentence, we first transform each sentence into successive window (n-gram) features to help disentangle factors of variations within sentences.
We choose sequence-based input other than relying on the syntactic parse trees before feeding in the neural network, thus our model doesn't rely on any external language knowledge and complicated pre-processing.\\
\indent In our experiments, we evaluate the semantic sentence
representations learned from C-LSTM with two tasks: sentiment
classification and 6-way question classification. Our evaluations show
that the C-LSTM model can achieve excellent results with several benchmarks
as compared with a wide range of baseline models. We also show that the
combination of CNN and LSTM outperforms individual multi-layer CNN
models and RNN models, which indicates that LSTM can learn
long-term dependencies from sequences of higher-level representations
better than the other models.

\section{Related Work}
Deep learning based neural network models have achieved great success in
many NLP tasks, including learning distributed word, sentence
and document representation~\cite{wordrep,pv}, parsing~\cite{parsing},
statistical machine translation~\cite{smt}, sentiment
classification~\cite{kim}, etc.
Learning distributed sentence representation through neural network
models requires little external domain knowledge and can reach
satisfactory results in related tasks like sentiment classification,
text categorization.\\
\indent In many recent sentence representation learning works, neural
network models are constructed upon either the input word sequences or
the transformed syntactic parse tree. Among them,
convolutional neural network (CNN) and recurrent neural network (RNN)
are two popular ones. \\
\indent The capability of capturing local correlations along with
extracting higher-level correlations through pooling empowers CNN to
model sentences naturally from consecutive context windows. In
~\cite{scratch}, Collobert et al. applied convolutional filters to
successive windows for a given sequence to extract global features by
max-pooling. As a slight variant, Kim et al.~\shortcite{kim} proposed a
CNN architecture with multiple filters (with a varying window size) and
two `channels' of word vectors. To capture word relations of varying
sizes, Kalchbrenner et al.~\shortcite{dcnn} proposed a dynamic k-max
pooling mechanism. In a more recent work~\cite{tao}, Tao et al. apply
tensor-based operations between words to replace linear operations on
concatenated word vectors in the standard convolutional layer and 
explore the non-linear interactions between nonconsective n-grams. Mou
et al.~\shortcite{mou} also explores convolutional models on
tree-structured sentences. \\
\indent As a sequence model, RNN is able to deal with variable-length
input sequences and discover long-term dependencies. Various variants of
RNN have been proposed to better store and access
memories~\cite{lstm,gru}. With the ability of explicitly modeling
time-series data, RNNs are being increasingly applied to sentence
modeling. For example, Tai et al.~\shortcite{tai2015} adjusted the
standard LSTM to tree-structured topologies and obtained superior results
over a sequential LSTM on related tasks.\\
\indent In this paper, we stack CNN and LSTM in a unified
architecture for semantic sentence modeling. The combination of CNN and
LSTM can be seen in some computer vision tasks like image
caption~\cite{caption} and speech recognition~\cite{saina}. Most of
these models use multi-layer CNNs and train CNNs and RNNs separately or
throw the output of a fully connected layer of CNN into RNN as inputs.
Our approach is different: we apply CNN to text data and feed
consecutive window features directly to LSTM, and so our architecture
enables LSTM to
learn long-range dependencies from higher-order sequential features. In
~\cite{lijiwei}, the authors suggest that sequence-based models are
sufficient to capture the compositional semantics for many NLP tasks,
thus in this work the CNN is directly built upon word sequences other
than the syntactic parse tree. Our experiments on sentiment
classification and 6-way question classification tasks clearly
demonstrate the
superiority of our model over single CNN or LSTM model and other related
sequence-based models.
\begin{figure}[t]
\centering
\includegraphics[width=3in]{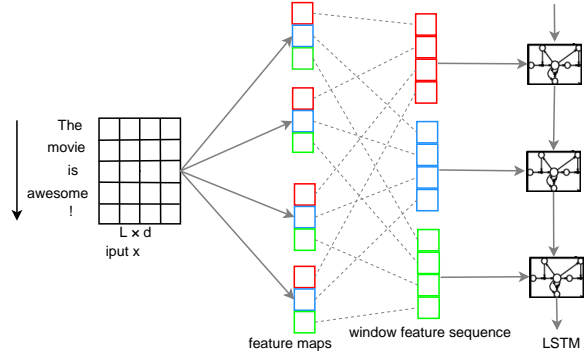}
\caption{The architecture of C-LSTM for sentence modeling. Blocks of the
same color in the feature map layer and window feature sequence layer
corresponds to features for the same window. The dashed lines connect
the feature of a window with the source feature map.  The final output
of the entire model is the last hidden unit of LSTM.}
\end{figure}

\section{C-LSTM Model}
The architecture of the C-LSTM model is shown in Figure 1, which
consists of two main components: convolutional neural network (CNN) and long
short-term memory network (LSTM). The following two subsections
describe how we apply CNN to extract higher-level sequences of word
features and LSTM to capture long-term dependencies over window feature
sequences respectively.

\subsection{N-gram Feature Extraction through Convolution}
The one-dimensional convolution involves a filter vector sliding over a
sequence and detecting features at different positions. Let
$\mathbf{x}_i \in \mathbb{R}^d$ be the $d$-dimensional word vectors for
the $i$-th word in a sentence. Let $\mathbf{x} \in \mathbb{R}^{L \times
d}$ denote the input sentence where $L$ is the length of the
sentence. Let $k$ be the length of the filter, and the vector
$\mathbf{m} \in \mathbb{R}^{k \times d}$ is a filter for the convolution
operation. For each position $j$ in the sentence, we have a window
vector $\mathbf{w}_j$ with $k$ consecutive word vectors, denoted as:
\begin{align}
\mathbf{w}_j = [\mathbf{x}_j,\mathbf{x}_{j+1}, \cdots ,\mathbf{x}_{j+k-1}]
\end{align}
Here, the commas represent row vector concatenation. A filter
$\mathbf{m}$ convolves with the window vectors ($\mathbf{k}$-grams) at each
position in a valid way to generate a feature map $\mathbf{c} \in
\mathbb{R}^{L-k+1}$; each element $c_j$ of the feature map for window
vector $\mathbf{w}_j$ is produced as follows:
\begin{align}
c_j = f(\mathbf{w}_j \circ \mathbf{m} + b),
\end{align}
where $\circ$ is element-wise multiplication, $b \in \mathbb{R}$ is a
bias term and $f$ is a nonlinear transformation function that can be
sigmoid, hyperbolic tangent, etc. In our case, we choose
ReLU~\cite{relu} as the nonlinear function. The C-LSTM model uses
multiple filters to generate multiple feature maps. For $n$ filters with
the same length, the generated $n$ feature maps can be rearranged as
feature representations for each window $w_j$,
\begin{align}
\mathbf{W} = [\mathbf{c}_1;\mathbf{c}_2; \cdots;\mathbf{c}_n]
\end{align}
Here, semicolons represent column vector concatenation and
$\mathbf{c}_i$ is the feature map generated with the $i$-th filter. Each
row $\mathbf{W}_j$ of $\mathbf{W} \in \mathbb{R}^ {(L-k+1)\times n}$ is
the new feature representation generated from $n$ filters for the window
vector at position $j$. The new successive higher-order window
representations then are fed into LSTM which is described below.\\
\indent A max-over-pooling or dynamic k-max pooling is often applied to
feature maps after the convolution to select the most or the k-most
important features. However, LSTM is specified for sequence input, and
pooling will break such sequence organization due to the discontinuous
selected features. Since we stack an LSTM neural neural network on top of
the CNN, we will not apply pooling after the convolution operation.
\subsection{Long Short-Term Memory Networks}
Recurrent neural networks (RNNs) are able to propagate historical
information via a chain-like neural network architecture. While
processing sequential data, it looks at the current input $x_t$ as well
as the previous output of hidden state $h_{t-1}$ at each time step.
However, standard RNNs becomes unable to learn long-term dependencies as
the gap between two time steps becomes large.
To address this issue, LSTM was first introduced in ~\cite{lstm} and
re-emerged as a successful architecture since Ilya et
al.~\shortcite{seq} obtained remarkable performance in statistical
machine translation. Although many variants of LSTM were proposed, we
adopt the standard architecture~\cite{lstm} in this work.\\
\indent The LSTM architecture has a range of repeated modules for each
time step as in a standard RNN. At each time step, the output of the
module is controlled by a set of gates in $\mathbb{R}^d$ as a function
of the old hidden state $h_{t-1}$ and the input at the current time step
$x_t$: the forget gate $f_t$, the input gate $i_t$, and the output gate
$o_t$. These gates collectively decide how to update the current memory
cell $c_t$ and the current hidden state $h_t$. We use $d$ to denote
the memory dimension in the LSTM and all vectors in this architecture
share the same dimension.
The LSTM transition functions are defined as follows:
\begin{align}
& i_t = \sigma(W_i \cdot [h_{t-1}, x_t] + b_i)
\\ \nonumber & f_t = \sigma(W_f \cdot [h_{t-1}, x_t] + b_f)
\\ \nonumber & q_t = \tanh(W_q \cdot [h_{t-1},x_t] + b_q)
\\ \nonumber & o_t = \sigma(W_o \cdot [h_{t-1}, x_t] + b_o)
\\ \nonumber & c_t = f_t \odot c_{t-1} + i_t \odot q_t
\\ \nonumber & h_t = o_t \odot \tanh(c_t)
\end{align}
Here, $\sigma$ is the logistic sigmoid function that has an output in
$[0,1]$, $\tanh$ denotes the hyperbolic tangent function that has an
output in $[-1, 1]$, and $\odot$ denotes the elementwise multiplication.
To understand the mechanism behind the architecture, we can view
$f_t$ as the function to control to
what extent the information from the old memory cell is going to be
thrown away, $i_t$ to control how much new information is going to be
stored in the current memory cell, and $o_t$ to control what to output
based on the memory cell $c_t$. LSTM is explicitly designed for
time-series data for learning long-term dependencies, and therefore we choose
LSTM upon the convolution layer to learn such dependencies in the
sequence of higher-level features.
\section{Learning C-LSTM for Text Classification}
For text classification, we regard the output of the hidden state at the
last time step of LSTM as the document representation and we add a
softmax layer on top. We train the entire model by minimizing
the cross-entropy error. Given a training sample $\mathbf{x}^{(i)}$ and
its true label $y^{(i)} \in \{1,2, \cdots, k\}$ where $k$ is the number
of possible labels and the estimated probabilities
$\widetilde{y}_j^{(i)} \in [0,1]$ for each label $j \in
\{1,2,\cdots,k\}$, the error is defined as:
\begin{align}
L(\mathbf{x}^{(i)}, y^{(i)}) = \sum_{j=1}^{k}{1\{y^{(i)} = j
\}\log(\widetilde{y}_j^{(i)})},
\end{align}
where $1\{\text{condition}\}$ is an indicator such that $1\{\text{condition is
true}\} = 1$ otherwise $1\{\text{condition is false}\}=0$.
We employ stochastic gradient descent (SGD) to learn the model
parameters and adopt the optimizer RMSprop~\cite{rmsprop}.

\subsection{Padding and Word Vector Initialization} First, we use
$maxlen$ to denote the maximum length of the sentence in the training
set. As the
convolution layer in our model requires fixed-length input, we pad
each sentence that has a length less than $maxlen$ with special symbols
at the end that indicate the unknown words. For a sentence in the test
dataset, we pad sentences that are shorter than $maxlen$ in the same
way, but for sentences that have a length longer than $maxlen$, we
simply cut extra words at the end of these sentences to reach $maxlen$.\\
\indent We initialize word vectors with the publicly available
\texttt{word2vec} vectors\footnote{\url{http://code.google.com/p/word2vec/}}
that are pre-trained
using about 100B words from the Google News Dataset. The dimensionality of
the word vectors is 300. We also initialize the word vector for the unknown
words from the uniform distribution [-0.25, 0.25]. We then fine-tune the word
vectors along with other model parameters during training.

\subsection{Regularization}
For regularization, we employ two commonly used techniques:
dropout~\cite{dropout} and L2 weight regularization. We apply dropout to
prevent co-adaptation. In our model, we either apply dropout to word
vectors before feeding the sequence of words into the convolutional
layer or to the output of LSTM before the softmax layer. The L2
regularization is applied to the weight of the softmax layer.

\section{Experiments}
We evaluate the C-LSTM model on two tasks: (1) sentiment classification,
and (2) question type classification. In this section, we introduce the
datasets and the experimental settings.

\subsection{Datasets}
{\bf Sentiment Classification: } Our task in this regard is to
predict the sentiment polarity of movie reviews. We use the
Stanford Sentiment Treebank (SST) benchmark~\cite{socher2013}. This
dataset consists of 11855 movie reviews and are split into train (8544),
dev (1101), and test (2210). Sentences in this corpus are parsed and all
phrases along with the sentences are fully annotated with 5 labels: very
positive, positive, neural, negative, very negative. We consider two
classification tasks on this dataset: fine-grained classification with 5
labels and binary classification by removing neural labels. For the
binary classification, the dataset has a split of
train (6920) / dev (872) / test (1821). Since the data is provided in the
format of sub-sentences, we train the model on both phrases and sentences but
only test on the sentences as in several previous
works~\cite{socher2013,dcnn}.\\
{\bf Question type classification: }Question classification is an
important step in a question answering system that classifies a question
into a specific type, e.g. \textit{``what is the highest waterfall in
the United States?"} is a question that belongs to ``location''. For
this task, we use the benchmark TREC~\cite{trec}. TREC divides all
questions into 6 categories, including \texttt{location, human, entity,
abbreviation, description} and \texttt{numeric}. The training dataset
contains 5452 labelled questions while the testing dataset contains 500
questions.
\subsection{Experimental Settings}
We implement our model based on Theano \cite{theano} -- a python library, which supports efficient symbolic differentiation and transparent use of a GPU. 
To benefit from the efficiency of parallel computation of the tensors, we train the model on
a GPU. For text preprocessing, we only convert all characters in the
dataset to lower case.\\
\indent For SST, we conduct hyperparameter (number of filters, filter
length in CNN; memory dimension in LSTM; dropout rate and which layer to
apply, etc.) tuning on the validation data in the standard split. For
TREC, we hold out 1000 samples from the training dataset for
hyperparameter search and train the model using the remaining data.\\
\indent In our final settings, we only use one convolutional layer and
one LSTM layer for both tasks. For the filter size, we investigated
filter lengths of 2, 3 and 4 in two cases: a) single convolutional layer
with the same filter length, and b) multiple convolutional layers with
different lengths of filters in parallel. Here we denote the number of
filters of length $i$ by $n_i$ for ease of clarification. For the first
case, each n-gram window is transformed into $n_i$ convoluted features
after convolution and the sequence of window representations is fed into
LSTM. For the latter case, since the number of windows generated from
each convolution layer varies when the filter length varies (see $L-k+1$
below equation (3)), we cut the window sequence at the end based on the
maximum filter length that gives the shortest number of windows. Each
window is represented as the concatenation of outputs from different
convolutional layers. We also exploit different combinations of
different filter lengths. We further present experimental analysis of
the exploration on filter size later. According to the experiments, we
choose a single convolutional layer with filter length 3.\\
\indent For SST, the number of filters of length 3 is set to be 150 and
the memory dimension of LSTM is set to be 150, too. The word vector
layer and the LSTM layer are dropped out with a probability of 0.5. For
TREC, the number of filters is set to be 300 and the memory dimension is
set to be 300. The word vector layer and the LSTM layer are dropped out with
a probability of 0.5. We also add L2 regularization with a factor of
0.001 to the weights in the softmax layer for both tasks.
\begin{table*}[t]
\begin{center}
\begin{tabular}{l|c|c|l}
\hline
\bf Model & \bf Fine-grained (\%)  & \bf Binary (\%) & \bf Reported in\\
\hline
SVM       & 40.7             & 79.4                & \cite{socher2013}\\
NBoW      & 42.4             & 80.5                & \cite{dcnn}\\
Paragraph Vector  & 48.7     & 87.8                & \cite{pv}\\
\hline
RAE       & 43.2             & 82.4                & (Socher, Pennington, et al., 2011)\\
MV-RNN    & 44.4             & 82.9                & \cite{mvrnn}\\
RNTN      & 45.7             & 85.4                & \cite{socher2013}\\
DRNN              & 49.8     & 86.6                & \cite{drnn}\\
\hline
CNN-non-static    & 48.0     & 87.2                & \cite{kim}\\
CNN-multichannel  & 47.4     & 88.1                & \cite{kim}\\
DCNN     & 48.5              & 86.8                & \cite{dcnn}\\
Molding-CNN & 51.2           & 88.6                & \cite{tao}\\
\hline
Dependency Tree-LSTM & 48.4  & 85.7                & \cite{tai2015}\\
Constituency Tree-LSTM & 51.0& 88.0                & \cite{tai2015}\\
LSTM     & 46.6              & 86.6                & our implementation\\
Bi-LSTM  & 47.8              & 87.9                & our implementation\\
\hline
C-LSTM   & 49.2              & 87.8                & our implementation\\
\hline
\end{tabular}
\end{center}
\caption{\label{sst} Comparisons with baseline models on Stanford
Sentiment Treebank. {\bf Fine-grained} is a 5-class classification task.
{\bf Binary} is a 2-classification task. The second block contains
the recursive models. The third block are methods related to convolutional
neural networks. The fourth block contains methods using LSTM (the first
two methods in this block also use syntactic parsing trees). The first
block contains other baseline methods. The last block is our model.}
\end{table*}
\section{Results and Model Analysis}
In this section, we show our evaluation results on sentiment
classification and question type classification tasks. Moreover, we give
some model analysis on the filter size configuration.
\subsection{Sentiment Classification}
\begin{table*}
\begin{center}
\begin{tabular}{l|c|l}
\hline
\bf Model & \bf Acc & \bf Reported in\\
\hline
SVM       & 95.0            & Silva et al~.\shortcite{svmtrec}\\
\hline
Paragraph Vector & 91.8     & Zhao et al~.\shortcite{zhao}\\
\hline
Ada-CNN   & 92.4            & Zhao et al~.\shortcite{zhao}\\
\hline
CNN-non-static & 93.6       & Kim~\shortcite{kim}\\
\hline
CNN-multichannel & 92.2     & Kim~\shortcite{kim}\\
\hline
DCNN    & 93.0              & Kalchbrenner et al.~\shortcite{dcnn}\\
\hline
LSTM    & 93.2              & our implementation\\
\hline
Bi-LSTM & 93.0             & our implementation\\
\hline
C-LSTM   &  94.6           & our implementation\\
\hline
\end{tabular}
\end{center}
\caption{\label{trec} The 6-way question type classification accuracy on TREC. }
\end{table*}
The results are shown in Table 1. We compare our model with a large
set of well-performed models on the Stanford Sentiment Treebank.\\
\indent Generally, the baseline models consist of recursive models,
convolutional neural network
models, LSTM related models and others. The recursive models employ a
syntactic parse tree as the sentence structure
and the sentence representation is computed recursively in
a bottom-up manner along the parse tree. Under this category, we choose
recursive autoencoder ({\bf RAE}), matrix-vector
({\bf MV-RNN}), tensor based composition ({\bf RNTN}) and multi-layer stacked ({\bf DRNN})
recursive neural network as baselines. Among CNNs, we compare with
Kim's~\shortcite{kim} CNN model with fine-tuned word vectors
({\bf CNN-non-static}) and multi-channels ({\bf CNN-multichannel}), {\bf DCNN} with
dynamic k-max pooling, Tao's CNN ({\bf Molding-CNN}) with low-rank tensor
based non-linear and non-consecutive convolutions. Among LSTM related
models, we first compare
with two tree-structured LSTM models ({\bf Dependence Tree-LSTM} and
{\bf Constituency Tree-LSTM}) that adjust LSTM to
tree-structured network topologies. Then we implement one-layer LSTM and
Bi-LSTM by ourselves. Since we could not tune the result of Bi-LSTM to
be as good as what has been reported in~\cite{tai2015} even if following
their untied weight configuration, we report our own results. For other
baseline methods, we compare against {\bf SVM} with unigram and bigram
features, {\bf NBoW} with average word vector features and {\bf paragraph vector}
that infers the new paragraph vector for unseen documents.\\
\indent To the best of our knowledge, we achieve the fourth best
published result for the 5-class classification task on this dataset.
For the binary classification task, we achieve comparable results with
respect to the state-of-the-art ones.
From Table 1, we have the following observations: (1) Although we did not beat the state-of-the-art ones, as an end-to-end model, the result is still promising and comparable with thoes models that heavily rely on linguistic annotations and knowledge, especially syntactic parse trees. This indicates C-LSTM will be more feasible for various scenarios.
(2) Comparing our results against single CNN and LSTM models shows that LSTM does learn long-term dependencies across sequences of higher-level representations better. We could explore in
the future how to learn more compact higher-level representations by
replacing standard convolution with other non-linear feature mapping
functions or appealing to tree-structured topologies before the
convolutional layer.

\subsection{Question Type Classification}
The prediction accuracy on TREC question classification is reported in
Table 2. We compare our model with a variety of models. The {\bf SVM}
classifier uses unigrams, bigrams, wh-word, head word, POS tags,
parser, hypernyms, WordNet synsets as engineered features and 60
hand-coded rules. {\bf Ada-CNN} is a self-adaptiive hierarchical sentence
model with gating networks. Other baseline models have been introduced
in the last task. From Table 2, we have the following observations:
(1) Our result consistently outperforms all published neural baseline models, which means that C-LSTM
captures intentions of TREC questions well.
(2) Our result is close to that of the state-of-the-art SVM that depends on highly engineered features. Such engineered features not only demands human laboring but also leads to the error propagation in the existing NLP tools, thus couldn't generalize well in other datasets and tasks. With the ability of automatically learning semantic sentence representations, C-LSTM doesn't require any human-designed features and has a better scalibility.

\begin{figure}[th]
\centering
\includegraphics[width=3.5in]{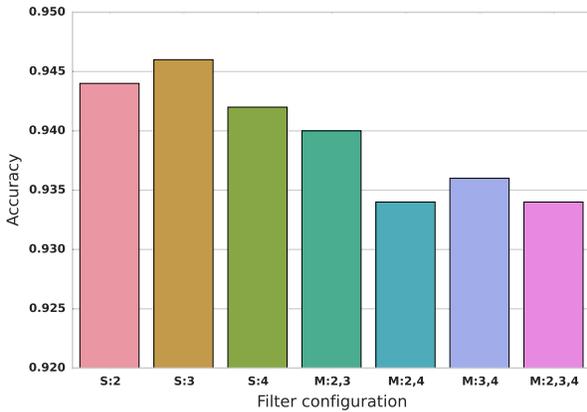}
\caption{Prediction accuracies on TREC questions with different filter
size strategies. For the horizontal axis, S means single convolutional
layer with the same filter length, and M means multiple convolutional
layers in parallel with different filter lengths.}
\end{figure}

\subsection{Model Analysis}
Here we investigate the impact of different filter
configurations in the convolutional layer on the model performance. \\
\indent In the convolutional layer of our model, filters are used to
capture local n-gram features. Intuitively, multiple convolutional
layers in parallel with different filter sizes should perform better
than single convolutional layers with the same length filters in that
different filter sizes could exploit features of different n-grams.
However, we found in our experiments that single convolutional layer
with filter length 3 always outperforms the other cases.\\
\indent We show in Figure 2 the prediction accuracies on the 6-way
question classification task using different filter configurations. Note
that we also observe the similar phenomenon in the sentiment
classification task. For each filter configuration, we report in
Figure 2 the best result under extensive grid-search on hyperparameters.
It it shown that single convolutional layer with filter length 3 performs
best among all filter configurations. For the case of multiple
convolutional layers in parallel, it is shown that filter configurations
with filter length 3 performs better that those without tri-gram
filters, which further confirms that tri-gram features do play a
significant role in capturing local features in our tasks. We conjecture
that LSTM could learn better semantic sentence representations from
sequences of tri-gram features.




\section{Conclusion and Future Work}
We have described a novel, unified model called C-LSTM that combines
convolutional neural network with long short-term memory network (LSTM).
C-LSTM is able to learn phrase-level features through a
convolutional layer; sequences of such higher-level
representations are then fed into the LSTM to learn long-term dependencies. We
evaluated the learned semantic sentence representations on sentiment
classification and question type classification tasks with very
satisfactory results.\\
\indent We could explore in the future ways to replace the standard
convolution with tensor-based operations or tree-structured convolutions.
We believe LSTM will benefit from more structured higher-level
representations.

\bibliography{clstm}
\bibliographystyle{naaclhlt2016}

\end{document}